\documentclass{article}
\usepackage{ijcai26}

\usepackage{times}
\usepackage{soul}
\usepackage{url}
\usepackage[hidelinks]{hyperref}
\usepackage[utf8]{inputenc}
\usepackage[small]{caption}
\usepackage{graphicx}
\usepackage{amsmath}
\usepackage{amsthm}
\usepackage{booktabs}
\usepackage{algorithm}
\usepackage{algorithmic}
\usepackage[switch]{lineno}

\usepackage{multirow}
\usepackage{amsfonts} 
\usepackage[table]{xcolor}
\usepackage{colortbl}
\usepackage{arydshln}
\usepackage{bm} 
\definecolor{tblHeader}{RGB}{32,55,118}
\definecolor{tblSubHdr}{RGB}{226,232,240}
\definecolor{tblOdd}{RGB}{248,250,252}
\definecolor{tblEven}{RGB}{241,245,249}
\definecolor{bestCell}{RGB}{251,207,232}    
\definecolor{secondCell}{RGB}{226,232,240}  

\newcommand{\pmv}[2]{\ensuremath{#1\!\pm\!#2}}
\newcommand{\best}[1]{\cellcolor{bestCell}{$\bm{#1}$}}
\newcommand{\second}[1]{\cellcolor{secondCell}{$\bm{#1}$}}

\title{Rethinking Drug–Drug Interaction Modeling as Generalizable Relation Learning}

\author{
Dong Xu$^1$\textsuperscript{*}
\and
Jiantao Wu$^1$\textsuperscript{*}
\and
Qihua Pan$^1$
\and
Sisi Yuan$^1$
\and
Zexuan Zhu$^1$
\and
Junkai Ji$^1$\textsuperscript{\dag}\\
\affiliations
$^1$School of Artificial Intelligence, Shenzhen University, Shenzhen 518060, China.\\
\emails
2400671001@mails.szu.edu.cn,
jonty395958@gmail.com,
jijunkai@szu.edu.cn\\
{\footnotesize
\textsuperscript{*}Dong Xu and Jiantao Wu contributed equally to this work.\\
\textsuperscript{\dag}Corresponding author: Junkai Ji.\\
}
}

\begin{document}

\maketitle

\begin{abstract}
Drug–drug interaction (DDI) prediction is central to drug discovery and clinical development, particularly in the context of increasingly prevalent polypharmacy. Although existing computational methods achieve strong performance on standard benchmarks, they often fail to generalize to realistic deployment scenarios, where most candidate drug pairs involve previously unseen drugs and validated interactions are scarce. We demonstrate that proximity in the embedding spaces of prevailing molecule-centric DDI models does not reliably correspond to interaction labels, and that simply scaling up model capacity therefore fails to improve generalization. To address these limitations, we propose GenRel-DDI, a generalizable relation learning framework that reformulates DDI prediction as a relation-centric learning problem, in which interaction representations are learned independently of drug identities. This relation-level abstraction enables the capture of transferable interaction patterns that generalize to unseen drugs and novel drug pairs. Extensive experiments across multiple benchmark demonstrate that GenRel-DDI consistently and significantly outperforms state-of-the-art methods, with particularly large gains on strict entity-disjoint evaluations, highlighting the effectiveness and practical utility of relation learning for robust DDI prediction. The code is available at \url{https://github.com/SZU-ADDG/GenRel-DDI}

\end{abstract}

\section{Introduction}
Drug–drug interaction (DDI) prediction constitutes a foundational task in drug discovery and clinical development \cite{DDI-Clinical}. It plays a central role in early safety assessment, therapeutic regimen prioritization, and the rational design of combination therapies, as DDIs can substantially alter drug exposure, efficacy, and toxicity profiles \cite{AAPS-DDI-effect}. Modern clinical practice is increasingly dominated by polypharmacy, resulting in a vast and continually expanding space of potential drug combinations that cannot be exhaustively evaluated through clinical trials alone \cite{Cell-combination,polypharmacy-dataset}.

Despite achieving strong performance on established benchmarks, existing computational approaches for DDI prediction often exhibit limited generalization to previously unseen data \cite{DDI-Ben}. In real-world deployment scenarios, the vast majority of candidate drug pairs involve drugs absent from the training set, posing a significant challenge to model robustness and reliability \cite{li2025identify}. Moreover, because experimentally or clinically validated DDIs remain scarce, DDI prediction inherently suffers from severe class imbalance, which biases model optimization and substantially reduces sensitivity to true interactions \cite{NIPS-lascal,ulichney2025double}. To more rigorously evaluate DDI models, a principled benchmark is proposed to explicitly control drug overlap between training and test sets, including unseen-pair, unseen-drug and strict entity-disjoint splits \cite{cold-start}.

\begin{figure}[t]
  \centering
  \includegraphics[width=0.86\linewidth]{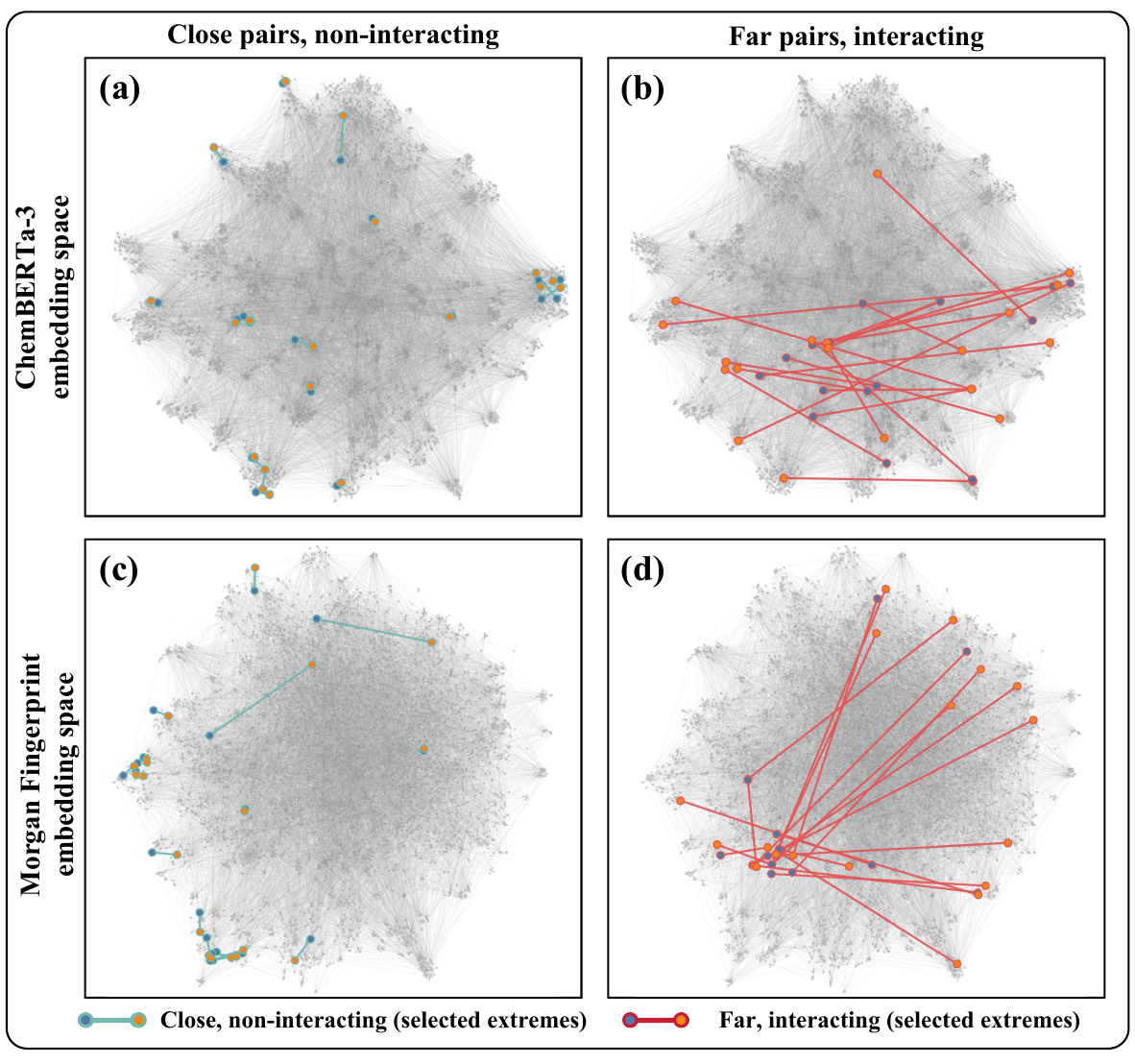}
  \vspace{-2mm}
  \caption{Proximity does not determine DDI labels. Closest pairs are not non-interacting (cyan), while farthest pairs still exhibit interactions (red) in both ChemBERTa-3 embedding and Morgan fingerprint spaces.}
  \label{fig:embeddings_space}
  \vspace{-4mm}
\end{figure}

\begin{figure*}[htbp]
  \centering
  \includegraphics[width=0.8\linewidth]{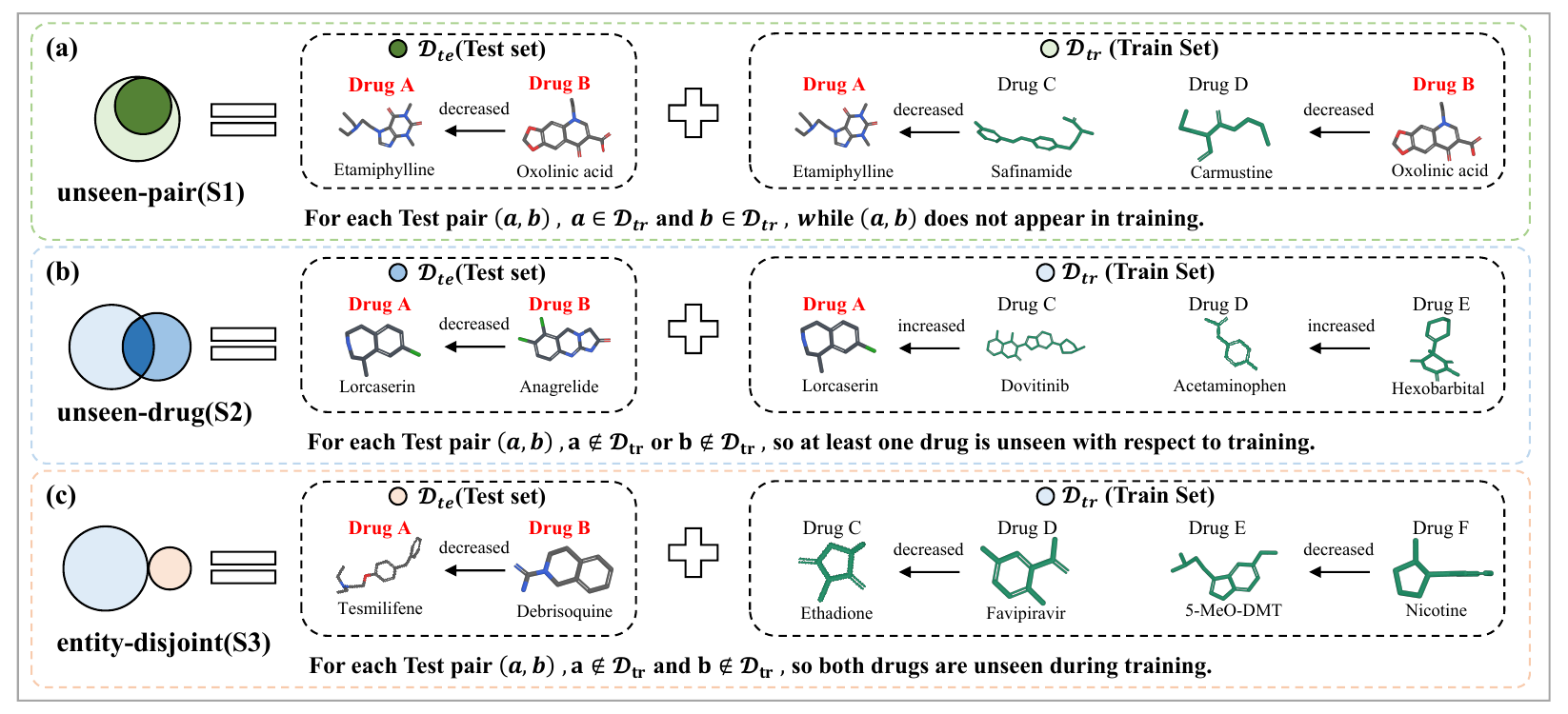}
  \vspace{-2mm}
  \caption{Illustration of evaluation splits by train--test drug overlap. (a) Unseen-pair (S1): both drugs are seen in training ($a,b\in\mathcal{V}_{\mathrm{tr}}$) but the test pair is held out. (b) Unseen-drug (S2): exactly one drug is seen ($a\in\mathcal{V}_{\mathrm{tr}}$ XOR $b\in\mathcal{V}_{\mathrm{tr}}$). (c) Entity-disjoint (S3): neither drug is seen ($a,b\notin\mathcal{V}_{\mathrm{tr}}$). Real drug examples are shown for illustration.}
  \label{fig:motivate}
  \vspace{-2mm}
\end{figure*}

Most current DDI models adopt a molecule-centric pipeline, in which a molecular encoder, trained from scratch or initialized from pretraining, is used to derive representations for individual drugs, followed by an interaction head that combines paired embeddings via similarity-based operators or neural fusion modules to produce interaction predictions \cite{DDI-DL-based}. However, such architectures do not reliably achieve robust performance, especially under entity-disjoint evaluation settings \cite{ahmad2025systematic}. As illustrated in Figure \ref{fig:embeddings_space},  the closest drug pairs do not necessarily correspond to non-interacting pairs, while the farthest pairs may still exhibit interactions in both the ChemBERTa-3 embedding and the Morgan fingerprint spaces. 

To address this issue, we propose GenRel-DDI, a generalizable relation learning framework for DDI prediction. GenRel-DDI reformulates DDI modeling as a relation learning problem, in which interaction representations are learned independently of specific drug identities. 
This design enforces relation-level abstraction and enables the model to capture transferable interaction patterns, thereby substantially enhancing generalization capability. 
Extensive experiments demonstrate that GenRel-DDI consistently achieves superior performance across multiple benchmark, significantly outperforming existing state-of-the-art methods. Notably, GenRel-DDI exhibits particularly large performance gains on the strict entity-disjoint benchmarks, where generalization is most challenging. Collectively, these results validate both the effectiveness and practical applicability of GenRel-DDI, and highlight the merit of relation learning paradigm.

To sum up, our contributions are threefold.
\begin{itemize}
\item We provide empirical evidence that simply scaling up model capacity does not improve the generalization performance of DDI prediction models.
\item We introduce GenRel-DDI, a generalizable relation learning framework that reframes DDI prediction from a molecule-centric to a relation-centric paradigm.
\item Extensive experiments show that GenRel-DDI significantly outperforms state-of-the-art methods across multiple benchmarks.
\end{itemize}

\section{Related Work}
\label{sec:related_work}

\paragraph{Molecule-centric}
Early DDI prediction relied on handcrafted drug descriptors and classical supervised learners.
Early deep models learned molecular representations and predicted interaction labels, as in DeepDDI \cite{DeepDDI-Dataset}.
Many methods remain molecule-centric: each drug is encoded independently, and a learned pair module predicts the interaction; variants mainly differ in encoder backbones and interaction operators.
Graph neural networks are a common encoder choice, and multi-resolution message passing remains a strong baseline \cite{MR-GNN}.
Recent models augment molecular encoders with attention-based pair modules or motif-level structure \cite{Taco-DDI}.
External knowledge is also incorporated, for example via task-specific knowledge subgraphs \cite{KnowDDI} or local reasoning over biomedical knowledge graphs \cite{TK-DDI}.

\paragraph{Evaluation and generalization}
Performance is often strong with train--test drug overlap, but it drops under overlap restriction with limited or no entity overlap \cite{NMI-MeTDDI}.
This gap grows under sparse and imbalanced supervision; many candidate pairs are negative or unlabeled.
Recent benchmarks use overlap-controlled splits, such as unseen-pair and entity-disjoint evaluation, to test robustness beyond train--test overlap \cite{li2025identify}.
Generalization is also challenged by cross-source transfer and label-space shifts.
Representation learning also studies objectives that reduce spurious cues, such as contrastive learning \cite{CGIB} and information-bottleneck variants \cite{IIB-DDI}.

\begin{figure*}[htbp]
  \centering
  \includegraphics[width=\linewidth]{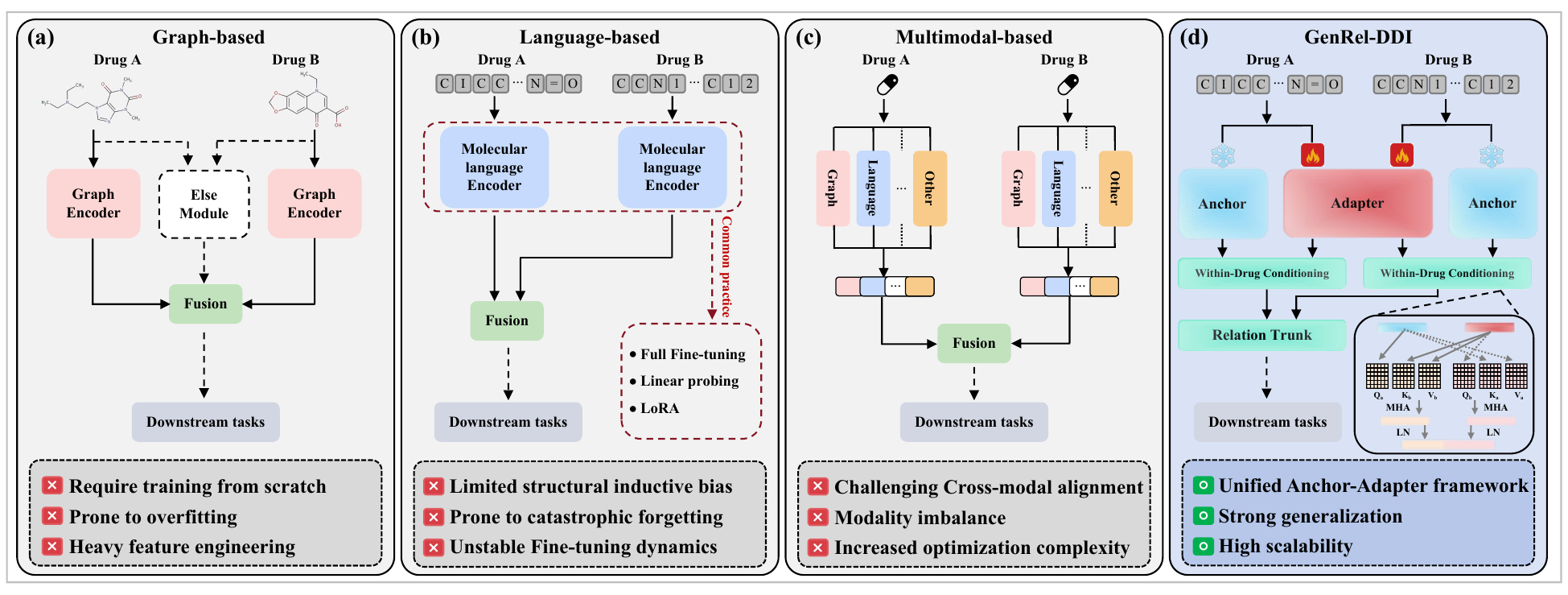}
  \vspace{-2mm}
  \caption{Overview of GenRel-DDI. Panels (a–c) illustrate common molecule-centric DDI pipelines that encode drugs independently and fuse pair features. Panel (d) shows GenRel-DDI, which keeps a pretrained encoder as a fixed anchor and trains a lightweight relation trunk with cross-attention to model partner-conditioned relational factors, reusing the same trunk across tasks by swapping only the head.}
  \label{fig:method_overview}
  \vspace{-4mm}
\end{figure*}

\section{Problem Setup}
\label{sec:problem_setup}

\paragraph{Task Definition.}

DDI prediction is formulated as supervised learning on drug pairs.
Let $a$ and $b$ denote two drugs with molecular inputs $x_a$ and $x_b$.
Here, $x_a$ denotes the molecular input associated with drug identity $a$.
Each instance is a pair $(a,b)$ with a benchmark-defined label $y\in\mathcal{Y}$.
We denote the dataset as $\mathcal{D}$ with $N$ labeled instances:
\begin{equation}
\mathcal{D}=\{((a_i,b_i),y_i)\}_{i=1}^N,\quad y_i\in\mathcal{Y}.
\end{equation}
A model predicts class probabilities $\hat{p}(y\mid x_a,x_b)$ over $\mathcal{Y}$ and is trained with standard cross-entropy on $\mathcal{D}_{\mathrm{tr}}$.

We consider two common label spaces.
Binary classification uses $\mathcal{Y}=\{0,1\}$.
Multi-class classification uses $\mathcal{Y}=\{1,\ldots,C\}$, where $C$ is benchmark-specific.

Benchmarks further differ in whether labels are \emph{directed} or \emph{undirected}.
For directed label spaces, $(a,b)$ and $(b,a)$ are treated as distinct instances and may carry different labels.
For undirected label spaces, a drug pair is treated as unordered and shares one label; duplicate unordered pairs are removed according to the benchmark convention.

\paragraph{Evaluation Splits by Train--Test Drug Overlap.}

Let $\mathcal{D}_{\mathrm{tr}}$ and $\mathcal{D}_{\mathrm{te}}$ denote the training and test sets.
We define the set of drug identities appearing in training as
\begin{equation}
\begin{aligned}
\mathcal{V}_{\mathrm{tr}}
&=
\{a \mid \exists\, b,y:((a,b),y)\in \mathcal{D}_{\mathrm{tr}}\}
\\
&\quad\cup
\{b \mid \exists\, a,y:((a,b),y)\in \mathcal{D}_{\mathrm{tr}}\}.
\end{aligned}
\end{equation}

Evaluation splits are defined by the overlap between test drugs and $\mathcal{V}_{\mathrm{tr}}$.
For directed label spaces, S1 uses a strict held-out definition: neither $((a,b),y)$ nor $((b,a),y)$ appears in $\mathcal{D}_{\mathrm{tr}}$ for any $y\in\mathcal{Y}$.
For undirected benchmarks, pairs are treated as unordered and deduplicated according to the benchmark convention (Suppl~C).
Let $\mathcal{V}_{\mathrm{te}}$ denote the set of drug identities appearing in $\mathcal{D}_{\mathrm{te}}$, and let $\mathcal{V}=\mathcal{V}_{\mathrm{tr}}\cup\mathcal{V}_{\mathrm{te}}$.
Figure~\ref{fig:motivate} illustrates the three overlap-controlled splits (S1-S3).

\section{Method: GenRel-DDI}
\label{sec:method}

Figure~\ref{fig:method_overview} contrasts three molecule-centric DDI pipelines with GenRel-DDI.
Prior pipelines encode each drug independently and then fuse; GenRel-DDI uses anchor--adapter within-drug conditioning and a role-separated relation trunk.

\subsection{Pretrained Reference Encoders}
\label{sec:method_enc}

Let $\{E^{(m)}_{\phi_m}\}_{m=1}^{M}$ denote $M\ge 1$ molecular encoders, where $\phi_m$ are the parameters of the $m$-th encoder.
For each drug identity $a$, the $m$-th encoder takes a tokenized molecular input $x_a^{(m)}$ and a binary mask $\mathbf{m}_a^{(m)}$.
$\mathbf{m}_a^{(m)}$ indicates valid tokens in the length-$T_a^{(m)}$ sequence, and the encoder outputs a token sequence $H_a^{(m)}$ with per-token hidden size $d_m$:
\begin{equation}
\begin{aligned}
H_a^{(m)} = E^{(m)}_{\phi_m}\!\left(x_a^{(m)},\mathbf{m}_a^{(m)}\right),\ \mathbf{m}_a^{(m)} \in \{0,1\}^{T_a^{(m)}} .
\end{aligned}
\end{equation}
The same applies to drug $b$, yielding $H_b^{(m)}$ and $\mathbf{m}_b^{(m)}$.

To make heterogeneous encoders compatible with a shared predictor (Suppl.~E), each encoder stream is mapped into a shared hidden space of dimension $d$ using a learnable per-token projection $\Pi^{(m)}_{\theta_m}:\mathbb{R}^{d_m}\rightarrow\mathbb{R}^{d}$:
\begin{equation}
\begin{aligned}
\bar{H}_a^{(m)} = \Pi^{(m)}_{\theta_m}\!\left(H_a^{(m)}\right), \ \bar{H}_b^{(m)} = \Pi^{(m)}_{\theta_m}\!\left(H_b^{(m)}\right).
\end{aligned}
\end{equation}
Here, $\bar{H}_a^{(m)}$ denotes the projected token sequence of drug $a$ under encoder $m$.
In implementation, $\Pi^{(m)}_{\theta_m}$ is instantiated by a linear map followed by normalization.
Each encoder stream has an independent freezing indicator.
Let $\delta_m\in\{0,1\}$ indicate whether stream $m$ is frozen.
When $\delta_m=1$, the stream parameters $\eta_m=(\phi_m,\theta_m)$ are held fixed and gradients are not propagated through $E^{(m)}_{\phi_m}$ or $\Pi^{(m)}_{\theta_m}$.
Selective freezing confines representation drift on $\mathcal{V}_{\mathrm{te}}\setminus\mathcal{V}_{\mathrm{tr}}$ to trainable streams.

\paragraph{Prediction drift under selective freezing.}
Let $F_{\psi,\omega}$ denote the downstream predictor that takes the projected token sequences ${\bar{H}_a^{(m)}}{m=1}^{M}$ and ${\bar{H}_b^{(m)}}{m=1}^{M}$ and returns a predictive distribution $\hat{p}(\cdot\mid x_a,x_b)\in\Delta^{|\mathcal{Y}|}$.

\paragraph{Assumption 1 (Lipschitz in encoder token sequences).}
Assume $F_{\psi,\omega}$ is $L$-Lipschitz with respect to its encoder inputs in Frobenius norm:
for any two pairs of inputs $\{\bar{H}_a^{(m)},\bar{H}_b^{(m)}\}$ and $\{\bar{H}_a'^{(m)},\bar{H}_b'^{(m)}\}$,
\begin{equation}
\begin{aligned}
\big\|F_{\psi,\omega}(\bar{H}_a^{(m)},\bar{H}_b^{(m)})
-
F_{\psi,\omega}(\bar{H}_a'^{(m)},\bar{H}_b'^{(m)})\big\|_2
\\ \le
L\sum_{m=1}^{M}
\Big(\|\bar{H}_a^{(m)}-\bar{H}_a'^{(m)}\|_F
+
\|\bar{H}_b^{(m)}-\bar{H}_b'^{(m)}\|_F).
\end{aligned}
\end{equation}

\paragraph{Anchoring reduces prediction drift.}
Let $\{\eta_m^0\}_{m=1}^{M}$ denote pretrained stream parameters with
$\eta_m^0=(\phi_m^0,\theta_m^0)$, and let $\{\eta_m\}_{m=1}^{M}$ be the
parameters after training under a freeze pattern $\{\delta_m\}_{m=1}^{M}$,
where $\delta_m=1$ indicates that stream $m$ is frozen so that $\eta_m=\eta_m^0$.
Let $\mathcal{A}=\{m\mid \delta_m=0\}$ be the set of trainable (adaptive) streams.
Define the per-stream representation drift over
$\mathcal{V}=\mathcal{V}_{\mathrm{tr}}\cup\mathcal{V}_{\mathrm{te}}$ as
\begin{equation}
\begin{aligned}
\Delta_m
&=
\sup_{d\in\mathcal{V}}
\big\|
\bar{H}_d^{(m)}(\eta_m)
-
\bar{H}_d^{(m)}(\eta_m^0)
\big\|_F .
\end{aligned}
\end{equation}

Then, under Assumption 1, the prediction drift between the selective-freezing setting and its pretrained reference counterpart is bounded by
\begin{equation}
\begin{aligned}
&\sup_{(a,b)\in\mathcal{D}_{\mathrm{te}}}
\big\|
\hat{p}_{\{\eta_m\}}(\cdot\mid x_a,x_b)
- \\
&\hat{p}_{\{\eta_m^0\}}(\cdot\mid x_a,x_b)
\big\|_2 
\le
2L\sum_{m\in\mathcal{A}}\Delta_m .
\end{aligned}
\end{equation}
Tokenization is fixed per benchmark, so $\bar{H}_d^{(m)}(\cdot)$ is compared under the same input sequence for each drug identity $d$.
In particular, frozen streams satisfy $\Delta_m=0$ by construction, so the worst-case drift is confined to the adaptive subset.

Apply Assumption 1 to the two inputs induced by $\{\eta_m\}$ and $\{\eta_m^0\}$.
For any test pair $(a,b)$,
\begin{equation}\label{eq:drift_proof}
\begin{split}
&\big\|
\hat{p}_{\{\eta_m\}}(\cdot\mid x_a,x_b)-\hat{p}_{\{\eta_m^0\}}(\cdot\mid x_a,x_b)
\big\|_2
\\ 
&\le
L\sum_{m=1}^{M}
\Big(\|\bar{H}_a^{(m)}(\eta_m)-\bar{H}_a^{(m)}(\eta_m^0)\|_F
\\
&+
\|\bar{H}_b^{(m)}(\eta_m)-\bar{H}_b^{(m)}(\eta_m^0)\|_F\Big)
\le
2L\sum_{m=1}^{M}\Delta_m .
\end{split}
\end{equation}

If $\delta_m=1$, then $\eta_m=\eta_m^0$ and the corresponding term is zero, yielding
$2L\sum_{m\in\mathcal{A}}\Delta_m$.
Taking the supremum over $(a,b)\in\mathcal{D}_{\mathrm{te}}$ concludes the bound.

\subsection{Within-Drug Conditioning}
\label{sec:method_within}

Within-drug conditioning maps the two role-specific streams of each drug into a single drug-level representation.
For drug $a$, the input consists of a reference token sequence $(\bar{H}_a^{(r)},\mathbf{m}_a^{(r)})$
and an adaptive token sequence $(\bar{H}_a^{(t)},\mathbf{m}_a^{(t)})$.
The same construction applies to drug $b$.

We first define the cross-attention operator and masked mean pooling.
For a query sequence $Q$, key sequence $K$, value sequence $V$, and a key mask $M_K$, a conditioning block is
\begin{equation}
\begin{aligned}
&\mathrm{CA}(Q,K,V;M_K)
=
\mathrm{LayerNorm}\!\Big(
Q+\\
&\mathrm{Dropout}\!\big(\mathrm{MultiHeadAttn}(Q,K,V;M_K)\big)
\Big).
\end{aligned}
\end{equation}

For a token sequence $H$ and a binary mask $\mathbf{m}$,
\begin{equation}
\begin{aligned}
\mathrm{Pool}(H,\mathbf{m})
&=
\frac{\sum_{t}\mathbf{m}_t\,H_t}{\sum_{t}\mathbf{m}_t}.
\end{aligned}
\end{equation}

Using $\mathrm{CA}(\cdot)$ and $\mathrm{Pool}(\cdot)$, a directional conditioning kernel is defined.
Let $\alpha,\beta\in\{r,t\}$ with $\alpha\neq\beta$.
For a parameter set $\vartheta$, define $\Gamma_a(\alpha \leftarrow \beta;\vartheta)$ as
\begin{equation}
\mathrm{Pool}\!\Big(
\mathrm{CA}_{\vartheta}\!\big(
\bar{H}_a^{(\alpha)},\ \bar{H}_a^{(\beta)},\ \bar{H}_a^{(\beta)};\ \mathbf{m}_a^{(\beta)}
\big),
\ \mathbf{m}_a^{(\alpha)}
\Big).
\end{equation}

Based on these primitives, five fusion operators are considered: a concat-MLP baseline and four CA-based variants (two one-way directions, plus two two-way variants with untied vs.\ tied parameters).

A pooled concatenation baseline is defined as
\begin{flalign}
g_a
&=
\mathrm{MLP}\!\Big(
\big[
\mathrm{Pool}(\bar{H}_a^{(r)},\mathbf{m}_a^{(r)}),\ 
\mathrm{Pool}(\bar{H}_a^{(t)},\mathbf{m}_a^{(t)})
\big]
\Big).
\end{flalign}

Using $\Gamma_a(\cdot)$, two one-way CA operators are obtained by swapping the attention direction ($t\leftarrow r$ vs.\ $r\leftarrow t$).
Two two-way operators further aggregate both directions via concatenation and an MLP, differing only in whether the parameters are shared.
\begin{align}
g_a &= \Gamma_a(t \leftarrow r;\vartheta_{t\leftarrow r})\quad or
\quad g_a = \Gamma_a(r \leftarrow t;\vartheta_{r\leftarrow t}), \\
g_a &= \mathrm{MLP}\!\Big(\big[\Gamma_a(t \leftarrow r;\vartheta_{t\leftarrow r}),\ \Gamma_a(r \leftarrow t;\vartheta_{r\leftarrow t})\big]\Big), \label{eq:ca_twoway_untied} \\
g_a &= \mathrm{MLP}\!\Big(\big[\Gamma_a(t \leftarrow r;\vartheta),\ \Gamma_a(r \leftarrow t;\vartheta)\big]\Big). \label{eq:ca_twoway_tied}
\end{align}

\subsection{Role-Separated Relation Trunk}
\label{sec:method_trunk}

Partner-conditioned factors are extracted by a relation trunk operating on fused token sequences.
Let $\tilde{H}_a \in \mathbb{R}^{T_a \times d}$ and $\tilde{H}_b \in \mathbb{R}^{T_b \times d}$ denote the fused token sequences produced by the within-drug conditioning module, with binary masks $\mathbf{m}_a \in \{0,1\}^{T_a}$ and $\mathbf{m}_b \in \{0,1\}^{T_b}$ aligned to the token dimension.
Two directional interaction states are computed via cross-attention:
\begin{equation}
\resizebox{1.00\linewidth}{!}{$
\begin{aligned}
U_{a\mid b}
&= \mathrm{CA}\!\left(\tilde{H}_a,\tilde{H}_b,\tilde{H}_b;\ \mathbf{m}_b\right), 
U_{b\mid a}= \mathrm{CA}\!\left(\tilde{H}_b,\tilde{H}_a,\tilde{H}_a;\ \mathbf{m}_a\right).
\end{aligned}
$}
\label{eq:dir_ca}
\end{equation}

A pooled, relation-only representation is then formed:
\begin{equation}
\begin{aligned}
p_{a\mid b} &= \mathrm{Pool}\!\left(U_{a\mid b},\mathbf{m}_a\right),\  p_{b\mid a} = \mathrm{Pool}\!\left(U_{b\mid a},\mathbf{m}_b\right),
\end{aligned}
\label{eq:dir_pool}
\end{equation}
\begin{equation}
z_{ab} = [p_{a\mid b};\ p_{b\mid a}] \in \mathbb{R}^{2d}.
\label{eq:zab}
\end{equation}
For undirected benchmarks, $z_{ab}$ is computed directly on the de-duplicated pair provided by the benchmark, without additional re-ordering or symmetrization.

\begin{table*}[t]
\centering
\scriptsize
\setlength{\tabcolsep}{4.2pt}
\renewcommand{\arraystretch}{1.18}
\resizebox{\linewidth}{!}{%
\begin{tabular}{l !{\color{black!30}\vrule width 0.7pt} ccc !{\color{black!30}\vrule width 0.7pt} ccc !{\color{black!30}\vrule width 0.7pt} ccc !{\color{black!30}\vrule width 0.7pt} ccc}
\toprule
\rowcolor{tblHeader}
\textcolor{white}{\textbf{Model}} &
\multicolumn{3}{c}{\textcolor{white}{\textbf{Unseen-pair (S1)}}} &
\multicolumn{3}{c}{\textcolor{white}{\textbf{Unseen-drug (S2)}}} &
\multicolumn{3}{c}{\textcolor{white}{\textbf{Entity-disjoint (S3)}}} &
\multicolumn{3}{c}{\textcolor{white}{\textbf{DDInter}}} \\
\rowcolor{tblSubHdr}
& \textbf{ACC} & \textbf{AUROC} & \textbf{AUPR}
& \textbf{ACC} & \textbf{AUROC} & \textbf{AUPR}
& \textbf{ACC} & \textbf{AUROC} & \textbf{AUPR}
& \textbf{ACC} & \textbf{AUROC} & \textbf{AUPR} \\
\cmidrule(lr){2-4}\cmidrule(lr){5-7}\cmidrule(lr){8-10}\cmidrule(lr){11-13}

\rowcolors{1}{tblOdd}{tblEven}
DeepDDI \cite{DeepDDI-Dataset}  & \pmv{84.1}{0.2} & \pmv{97.1}{0.1} & \pmv{92.3}{0.1}
         & \pmv{66.4}{1.4} & \pmv{88.1}{1.1} & \pmv{72.0}{1.7}
         & \pmv{51.4}{2.9} & \pmv{78.1}{2.1} & \pmv{52.4}{2.6}
         & \pmv{48.9}{1.0} & \pmv{73.3}{0.5} & \pmv{42.4}{0.9} \\
DeepCCI \cite{DeepCCI}  & \pmv{89.8}{0.2} & \pmv{98.8}{0.1} & \pmv{96.6}{0.1}
         & \pmv{65.2}{1.3} & \pmv{87.8}{0.5} & \pmv{71.0}{1.3}
         & \pmv{50.3}{2.1} & \pmv{77.5}{1.3} & \pmv{50.1}{2.6}
         & \second{\pmv{58.9}{0.4}} & \pmv{82.0}{0.4} & \pmv{60.9}{0.6} \\
MR-GNN \cite{MR-GNN}   & \pmv{87.1}{0.2} & \pmv{98.1}{0.1} & \pmv{94.9}{0.1}
         & \pmv{65.3}{1.0} & \pmv{87.0}{0.8} & \pmv{68.9}{2.1}
         & \pmv{50.9}{2.8} & \pmv{75.7}{2.3} & \pmv{50.0}{3.5}
         & \pmv{48.2}{3.5} & \pmv{75.8}{1.1} & \pmv{49.4}{3.4} \\
SA-DDI \cite{SA-DDI}   & \pmv{91.8}{0.7} & \pmv{97.7}{3.1} & \second{\pmv{97.5}{0.3}}
         & \pmv{67.4}{0.8} & \pmv{88.4}{0.8} & \pmv{72.9}{1.3}
         & \pmv{52.4}{3.2} & \pmv{78.6}{2.3} & \pmv{53.6}{3.7}
         & \pmv{58.3}{0.7} & \pmv{81.8}{0.5} & \pmv{60.0}{0.7} \\
MolTrans \cite{MolTrans} & \pmv{87.6}{0.4} & \pmv{98.2}{0.1} & \pmv{95.0}{0.2}
         & \pmv{66.4}{0.8} & \pmv{88.2}{0.7} & \pmv{72.9}{0.6}
         & \pmv{48.8}{2.3} & \pmv{75.3}{3.0} & \pmv{48.4}{2.1}
         & \pmv{56.4}{2.2} & \pmv{81.0}{1.8} & \pmv{58.7}{3.0} \\
Molormer \cite{Molormer}  & \pmv{89.7}{1.6} & \pmv{98.7}{0.4} & \pmv{96.3}{1.0}
         & \pmv{66.7}{0.8} & \second{\pmv{88.6}{0.7}} & \second{\pmv{73.7}{1.1}}
         & \pmv{49.9}{2.9} & \pmv{77.7}{2.5} & \pmv{52.0}{3.5}
         & \pmv{58.8}{1.0} & \second{\pmv{82.2}{0.6}} & \pmv{61.1}{1.0} \\
CIGIN \cite{CIGIN}    & \pmv{87.4}{0.2} & \pmv{96.3}{0.1} & \pmv{90.3}{0.5}
         & \pmv{63.6}{2.1} & \pmv{86.1}{1.2} & \pmv{67.6}{2.2}
         & \pmv{51.1}{2.8} & \pmv{78.6}{2.1} & \pmv{52.9}{2.6}
         & \pmv{58.7}{0.9} & \pmv{82.1}{0.5} & \pmv{60.0}{0.9} \\
CGIB \cite{CGIB}     & \best{\pmv{97.3}{0.1}} & \pmv{98.8}{0.1} & \pmv{96.5}{0.3}
         & \pmv{67.5}{1.0} & \pmv{87.0}{0.8} & \pmv{69.5}{1.0}
         & \pmv{52.0}{3.3} & \pmv{78.5}{2.5} & \pmv{52.9}{3.1}
         & \pmv{56.8}{0.3} & \pmv{80.6}{0.4} & \pmv{59.0}{1.0} \\
MeTDDI \cite{NMI-MeTDDI}   & \pmv{92.3}{0.1} & \second{\pmv{99.2}{0}} & \best{\pmv{97.9}{0.1}}
         & \second{\pmv{68.0}{0.7}} & \pmv{88.5}{0.2} & \pmv{73.2}{1.7}
         & \second{\pmv{53.6}{2.2}} & \second{\pmv{78.9}{1.5}} & \second{\pmv{53.7}{1.9}}
         & \best{\pmv{59.2}{0.4}} & \best{\pmv{82.6}{0.5}} & \second{\pmv{61.8}{0.5}} \\
\specialrule{0.8pt}{2pt}{2pt}

GenRel-DDI ($C(*)+M$)
& \second{\pmv{92.5}{1.4}} & \best{\pmv{99.3}{0.3}} & \best{\pmv{97.9}{0.7}}
& \best{\pmv{68.8}{0.3}} & \best{\pmv{89.2}{0.1}} & \best{\pmv{75.1}{0.5}}
& \best{\pmv{55.1}{0.5}} & \best{\pmv{80.5}{0.5}} & \best{\pmv{57.6}{1.0}}
& \pmv{58.6}{1.4} & \best{\pmv{82.6}{1.3}} & \best{\pmv{62.5}{2.5}} \\
\bottomrule
\end{tabular}%
}
\caption{Main results on MeTDDI overlap splits and cross-source transfer evaluation on DDInter typed (directed, $C{=}4$).}
\vspace{-2mm}
\label{tab:ddi_splits_main}
\end{table*}

\begin{figure*}[htpb]
  \centering
  \includegraphics[width=\linewidth]{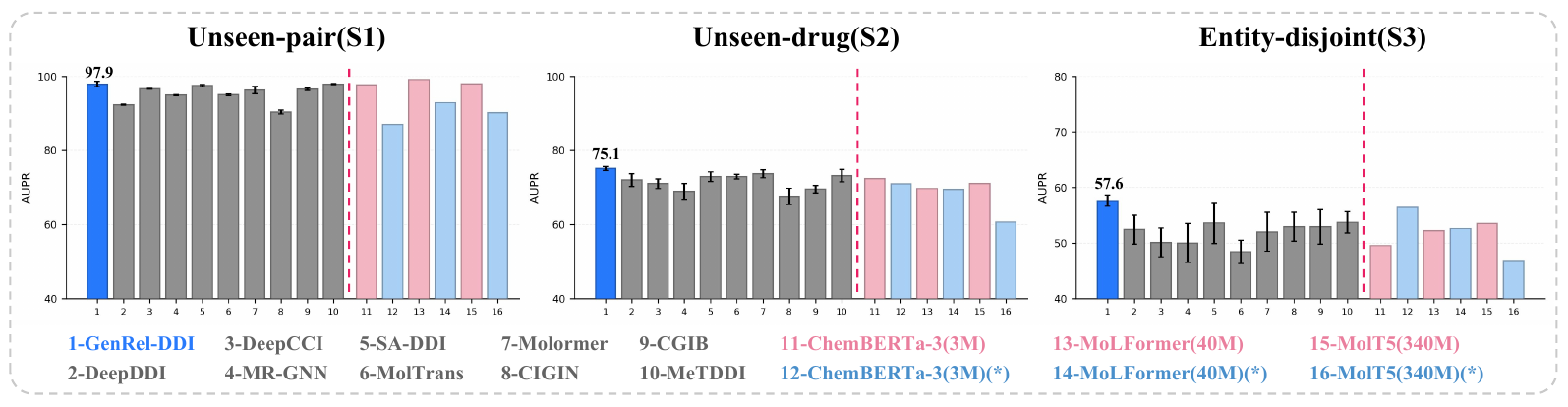}
  \vspace{-4mm}
  \caption{Scaling and tuning under overlap restriction on MeTDDI: AUPR on S1–S3 for three pretrained encoders under frozen and full
fine-tuning settings. Published baselines are shown for reference; $(\ast)$ indicates a frozen encoder. Error bars denote mean ± std over five runs.}
  \label{fig:overlap_splits_aupr}
  \vspace{-2mm}
\end{figure*}

\subsection{Prediction Head and Training}
\label{sec:method_head}

A task-specific head $g_{\omega}$ maps $z_{ab}$ to logits over $\mathcal{Y}$.
For binary classification with $\mathcal{Y}=\{0,1\}$,
\begin{equation}
\begin{aligned}
\hat{p}(y=1\mid x_a,x_b)
&= \sigma\!\left(g_{\omega}(z_{ab})\right),
\end{aligned}
\end{equation}
and for multi-class classification with $\mathcal{Y}=\{1,\ldots,C\}$,
\begin{equation}
\begin{aligned}
\hat{p}(y=c\mid x_a,x_b)
&= \mathrm{softmax}\!\left(g_{\omega}(z_{ab})\right)_c .
\end{aligned}
\end{equation}

Let $\mathcal{D}_{\mathrm{tr}}$ denote the training set defined in Section~\ref{sec:problem_setup}.
Cross-entropy is minimized:
\begin{equation}
\begin{aligned}
\mathcal{L}
&= -\sum_{((a,b),y)\in \mathcal{D}_{\mathrm{tr}}}
\log \hat{p}\!\left(y\mid x_a,x_b\right).
\end{aligned}
\end{equation}
The relation trunk, the task head, and encoder streams with $\delta_m=0$ are updated, while streams with $\delta_m=1$ remain fixed at their pretrained parameters.
This treats pretrained representations as a stable semantic anchor and allocates adaptive capacity to partner-conditioned relation learning under overlap restriction.

\section{Experiments}

\subsection{Overlap Splits and Transfer}
\label{sec:exp_group1}

\noindent\textbf{Setup.}
GenRel-DDI is evaluated on MeTDDI under three overlap splits (S1--S3) and tested on DDInter typed.
All numbers are averaged over five runs, and all metrics are micro-averaged.
For DDInter typed, the model trained on MeTDDI S3 is used without further training.
It is evaluated on the union of the official DDInter train and test pairs.
This follows prior cross-source evaluation practice and avoids target-domain tuning.
The MeTDDI dataset is a metabolic DDI benchmark derived from DrugBank \cite{DrugBank-5.0}.
Each record states whether a perpetrator increases or decreases the metabolism of a victim drug.
Following \cite{NMI-MeTDDI}, ordered pairs are used: $(a,b)$ and $(b,a)$ are distinct.
This yields four classes that combine direction and effect.
DDInter typed uses the same label space but draws interactions from DDInter \cite{DDInter}.
It is therefore a cross-source test set.

\paragraph{Baselines.}
Comparisons are made to methods commonly reported under the same S1--S3 splits: DeepDDI \cite{DeepDDI-Dataset}, DeepCCI \cite{DeepCCI}, MR-GNN \cite{MR-GNN}, SA-DDI \cite{SA-DDI}, MolTrans \cite{MolTrans}, Molormer \cite{Molormer}, CIGIN \cite{CIGIN}, CGIB \cite{CGIB}, and MeTDDI \cite{NMI-MeTDDI}.
Baseline numbers are taken from the original papers.

\begin{figure*}[!h]
  \centering
  \includegraphics[width=\linewidth]{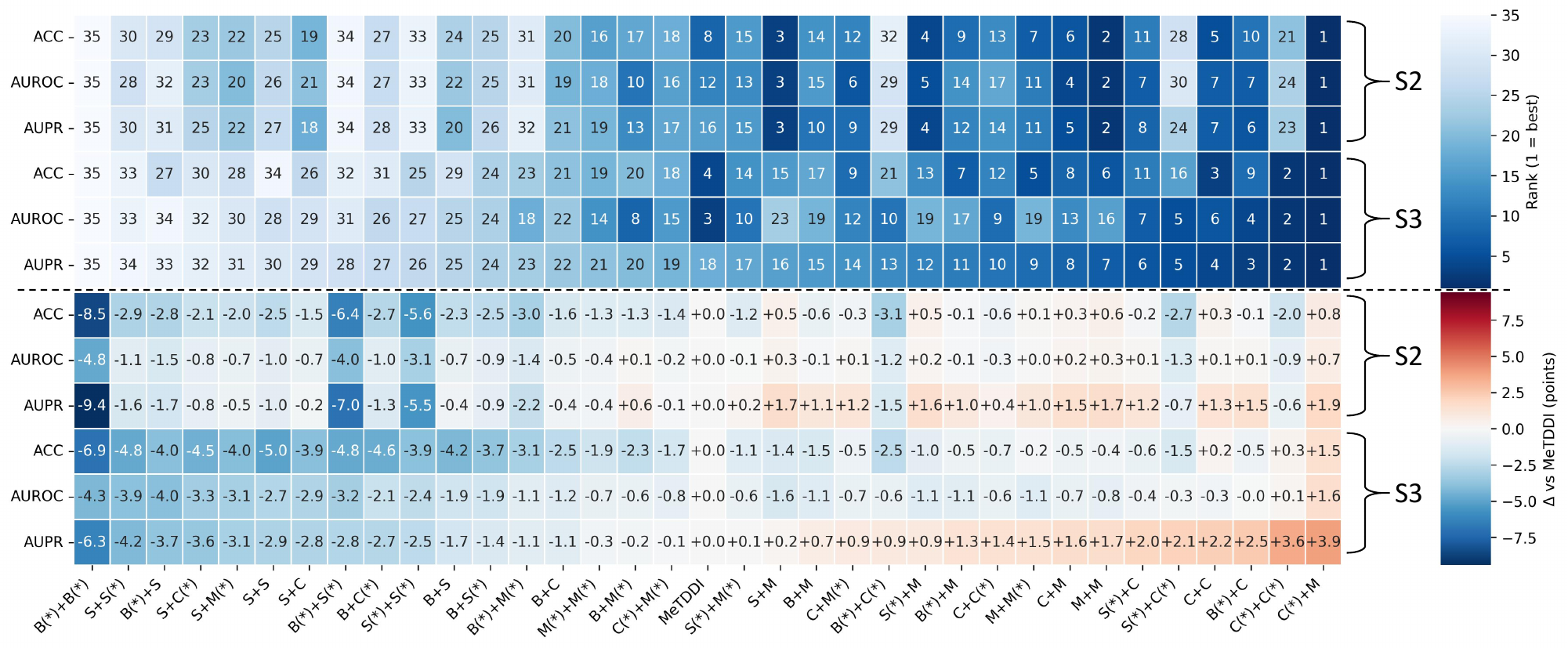}
  \vspace{-6mm}
  \caption{Role assignment and encoder pairing results of GenRel-DDI on MeTDDI under overlap-restricted splits (S2/S3). Each column is an ordered encoder pair $X{+}Y$, where $X$ is assigned to the anchor stream $(r)$ and $Y$ to the adapter stream $(t)$. $(*)$ indicates a frozen encoder stream. Top: per-metric ranks (1 = best) across all configurations on S2 and S3 for ACC/AUROC/AUPR. Bottom: absolute differences (points) relative to the MeTDDI baseline (the column labeled MeTDDI). Columns are sorted by S3 AUPR in ascending order.}
  \label{fig:role}
  \vspace{-2mm}
\end{figure*}

\paragraph{Results.}
Table~\ref{tab:ddi_splits_main} summarizes results on MeTDDI under S1--S3 and the cross-source test on DDInter typed.
Supplementary~B reports the ablations.
On S1, performance is close to ceiling, and GenRel-DDI reaches 99.3 AUROC.
On S2, GenRel-DDI ranks best on the three micro-averaged metrics.
On S3, GenRel-DDI is best on ACC, AUROC, and AUPR.
This result reflects stronger entity-disjoint generalization.
On DDInter typed, the S3-trained model matches MeTDDI on AUROC, improves AUPR, and remains close on ACC.
This supports cross-source generalization from MeTDDI to DDInter.
Across methods, performance drops from S1 to S3.
This confirms that overlap restriction is the main driver of generalization difficulty.

Supplementary~B examines the main design choices through ablations.
For stream composition, $C(\ast){+}M$ gives the strongest results on S3 and the cross-source test.
The \emph{Ref-only} variant remains strong on S2 and DDInter typed, whereas \emph{Ada-only} degrades on overlap-restricted settings.
This indicates that anchoring is needed under overlap restriction.
Changing the within-drug or within-pair operator leads to small changes without a consistent trend.
Most gains come from the anchor--adapter split and the role-separated relation trunk.
Under overlap restriction, encoder updates can shift representations on unseen drugs.
Freezing the anchor stream limits this drift.
This is consistent with Sec.~\ref{sec:method_enc}.
Freezing enforces $\Delta_m{=}0$ on anchored streams and confines drift to the adaptive subset, which is most relevant when many test drugs are unseen.
Encoder scaling and tuning are analyzed in Sec.~\ref{sec:exp_group2}.
Role assignment is analyzed in Sec.~\ref{sec:exp_group3}.

\begin{table*}[htbp]
\centering
\scriptsize
\setlength{\tabcolsep}{4.2pt}
\renewcommand{\arraystretch}{1.18}
\resizebox{\linewidth}{!}{%
\begin{tabular}{l !{\color{black!30}\vrule width 0.7pt} c c c c !{\color{black!30}\vrule width 0.7pt} c c c c !{\color{black!30}\vrule width 0.7pt} c c c c}
\toprule
\rowcolor{tblHeader}
\textcolor{white}{\textbf{Model}} &
\multicolumn{4}{c}{\textcolor{white}{\textbf{DeepDDI 86-class}}} &
\multicolumn{4}{c}{\textcolor{white}{\textbf{ZhangDDI Binary}}} &
\multicolumn{4}{c}{\textcolor{white}{\textbf{ChChDDI}}} \\
\rowcolor{tblSubHdr}
& \textbf{AUROC} & \textbf{AUPR} & \textbf{F1} & \textbf{ACC}
& \textbf{AUROC} & \textbf{AUPR} & \textbf{F1} & \textbf{ACC}
& \textbf{AUROC} & \textbf{AUPR} & \textbf{F1} & \textbf{ACC} \\
\cmidrule(lr){2-5}\cmidrule(lr){6-9}\cmidrule(lr){10-13}

\rowcolors{1}{tblOdd}{tblEven}
MR-GNN \cite{MR-GNN}   & 93.35 & 94.56 & 90.07 & 87.54 & 96.18 & 92.63 & 82.93 & 91.90 & 93.11 & 95.95 & 88.13 & 85.03 \\
EPGNN-DS \cite{EPGNN-DS} & 85.93 & 88.72 & 84.86 & 80.22 & 90.83 & 88.96 & 80.07 & 82.40 & 94.23 & 96.80 & 89.41 & 86.64 \\
DeepDrug \cite{DeepDrug} & 91.74 & 92.99 & 89.39 & 86.28 & 93.35 & 92.33 & 82.89 & 85.67 & 98.38 & 99.16 & 94.67 & 93.18 \\
MIRACLE \cite{MIRACLE}  & 92.76 & 96.77 & \ 93.54 & 90.33 & \second{96.44} & \second{93.09} & 85.16 & \second{93.16} & 96.20 & 99.50 & 94.55 & 90.77 \\
CSGNN \cite{CSGNN}    & 94.01 & 94.17 & 86.01 & 86.33 & 91.71 & 89.02 & 83.60 & 84.14 & 97.68 & 97.56 & 92.47 & 92.54 \\
SSI-DDI \cite{SSI-DDI}  & 91.79 & 93.47 & 88.23 & 85.38 & 93.14 & 92.09 & 81.96 & 85.35 & 98.09 & 98.97 & 93.98 & 92.19 \\
DeepDDS \cite{DeepDDS}  & 94.38 & 95.68 & 91.27 & 88.87 & 93.20 & 92.08 & 82.79 & 85.63 & 97.10 & 98.51 & 92.21 & 90.38 \\
DSN-DDI \cite{DSN-DDI}  & 93.22 & 92.87 & 85.60 & 85.41 & 91.13 & 86.42 & \second{87.68} & 86.65 & 96.69 & 96.34 & 88.12 & 88.89 \\
SMG-DDI \cite{SMG-DDI} & 76.76 & 83.32 & 79.11 & - & 90.49 & 90.04 & 81.29 & - & 78.97 & 88.18 & 85.11 & - \\
GoGNN \cite{Gognn} & 92.71 & - & 89.93 & 93.54 & 92.35 & - & 81.54 & 84.14 & 96.64 & - & 92.35 & 91.17 \\
MDF-SA-DDI \cite{MDF-SA-DDI} & 88.84 & - & 96.13 & 94.12 & 94.03 & - & 83.67 & 86.89 & 98.10 & - & 94.17 & \second{94.63} \\
CGIB \cite{CGIB} & 98.08 & - & \second{96.53} & 95.76 & 94.43 & - & 84.53 & 87.32 & 98.38 & - & \second{95.44} & 94.37 \\
IIB-DDI \cite{IIB-DDI} & 98.67 & - & \best{96.84} & 96.01 & 94.88 & - & 85.25 & 88.31 & 98.72 & - & \best{95.78} & \second{94.63} \\

\specialrule{1.2pt}{2pt}{2pt}

GenRel-DDI ($C(*){+}M$)   & \second{99.96} & \best{99.28} & 93.19 & \best{96.91} & \best{98.23} & \best{96.44} & \best{92.50} & \best{93.74} & \best{99.59} & \best{99.91} & 93.32 & \best{95.48} \\
GenRel-DDI ($C{+}M(*)$)   & \best{99.97} & \second{99.15} & 90.76 & \second{96.32} & 96.01 & 92.31 & 87.62 & 89.47 & \second{98.95} & \second{99.76} & 91.56 & 94.55 \\

\bottomrule
\end{tabular}%
}
\caption{
Evaluation across label spaces: DeepDDI, 86-class; ZhangDDI and ChChDDI, binary. Metrics are micro-averaged. A dash indicates an unreported metric in the original paper. Pink and gray mark the best and second-best value in each column among reported numbers, with ties highlighted consistently.
}
\vspace{-4mm}
\label{tab:three_datasets_single_table}
\end{table*}

\subsection{Scaling Study}
\label{sec:exp_group2}

\paragraph{Setup.}
Encoder scale and tuning strategy are examined under overlap-restricted splits.
Three pretrained molecular encoders are evaluated under two tuning modes: frozen and full fine-tuning.
All variants share the same GenRel-DDI relation trunk and head, and are trained and evaluated on MeTDDI under S1--S3.
To interpret encoder differences beyond parameter count, Supplementary~B visualizes pair spaces using kNN graphs and t-SNE.
The focus is the change from overlap-permitting S1 to entity-disjoint S3.
This comparison keeps the GenRel-DDI architecture fixed.

\paragraph{Results.}
Figure~\ref{fig:overlap_splits_aupr} shows AUPR on S1--S3 for the three encoders under frozen and full fine-tuning.
AUPR drops from S1 to S3 for all variants.
This indicates that entity-disjoint S3 is the hardest split.
On S1, full fine-tuning is consistently better than freezing.
Scaling adds little once S1 performance is already high.
On S2, the gap between full fine-tuning and freezing becomes smaller for most encoders.
On S3, full fine-tuning is not reliably better than freezing.
Freezing can match or exceed full fine-tuning, which is consistent with representation drift on unseen drugs.
With frozen encoders on S3, the ordering among ChemBERTa-3 \cite{chemberta-3}, MoLFormer\ \cite{NMI-Molformer}, and MolT5 \cite{MolT5} does not follow model size.
Scale alone therefore does not predict entity-disjoint generalization.
MolT5 behaves differently on S2 and S3 compared with the other encoders.
This suggests that the pretraining objective matters for downstream generalization.
After full fine-tuning, results are very close on S1 and remain close on S2.
Encoder choice is therefore less influential under overlap-permitting evaluation.

\subsection{Role Assignment and Encoder Pairing}
\label{sec:exp_group3}

\paragraph{Setup.}
Role assignment is evaluated by pairing two encoders in an ordered way and mapping them to the anchor stream $r$ and the adapter stream $t$.
Order matters because $r$ is frozen while $t$ is trainable in the default setting.
All settings follow Sec.~\ref{sec:exp_group1} and keep the relation trunk, head, and training hyperparameters fixed.
Only the encoder order and the frozen stream are changed.
Results are reported on S2 and S3 because role effects are most visible when at least one test drug is unseen.

\paragraph{Results.}
Strong anchor--adapter asymmetry is observed on S2 and S3.
$C(*){+}M$ is the best pairing.
It ranks first on ACC, AUROC, and AUPR on both S2 and S3, and improves over MeTDDI (e.g., $\Delta$AUPR $=+1.9/+3.9$ and $\Delta$AUROC $=+0.7/+1.6$ on S2/S3).
Freezing the adapter in the same pair reduces the gains.
For example, $C{+}M(*)$ drops on S3, where $\Delta$AUPR decreases from $+3.9$ to $+0.9$.
This suggests that partner-conditioned adaptation should be assigned to the adapter stream.
Heterogeneous pairings cannot be predicted from single-encoder performance.
Several settings increase S3 AUPR but reduce S3 ACC or AUROC.
This trade-off is expected under class imbalance, since AUPR is more sensitive to the minority class.
$C(*){+}M$ is the only setting that improves ACC, AUROC, and AUPR on S3 at the same time.
This points to a simple rule: keep the anchor stable and place trainable capacity in the adapter.
Figure~\ref{fig:role} shows that encoder order and role assignment are primary design choices.
Under overlap restriction, generalization depends more on the $r/t$ role split and where freezing is applied than on encoder scale.

\subsection{Evaluation Across Diverse Label Spaces}
\label{sec:exp_group4}

\paragraph{Setup.}
We evaluate GenRel-DDI on four benchmarks with different label spaces: ZhangDDI and ChChDDI, binary; DeepDDI, 86-class typed; and DDInter severity, 4-class severity.
ZhangDDI has 548 drugs and 48,548 labeled pairs.
ChChDDI has 1,514 drugs and 48,514 labeled pairs.
DeepDDI has 1,704 drugs and about 192k labeled pairs over 86 interaction types.
For DDInter severity, we use the processed subset released with MathEagle \cite{MathEagle}, with 1,603 drugs and 177,352 pairs.
Across benchmarks, the anchor--adapter backbone and the relation trunk are kept unchanged.
Using the same training setting as Sec.~\ref{sec:exp_group1}.
These benchmarks differ in supervision density and class imbalance.
This tests whether the same relation trunk transfers beyond dataset-specific signals.

\begin{table}[!h]
\centering
\scriptsize
\setlength{\tabcolsep}{4.2pt}
\renewcommand{\arraystretch}{1.18}
\resizebox{\linewidth}{!}{%
\begin{tabular}{l !{\color{black!30}\vrule width 0.7pt} c c c c c}
\toprule
\rowcolor{tblHeader}
\textcolor{white}{\textbf{Method}} &
\multicolumn{5}{c}{\textcolor{white}{\textbf{DDInter severity}}} \\
\rowcolor{tblSubHdr}
& \textbf{ACC} & \textbf{AUROC} & \textbf{AUPR} & \textbf{F1} & \textbf{MCC} \\
\cmidrule(lr){2-6}

\rowcolors{1}{tblOdd}{tblEven}
GraphSAGE \cite{GraphSAGE} & 83.16 & 93.19 & 88.10 & 83.16 & 75.74 \\
SSI-DDI \cite{SSI-DDI}    & 61.89 & 66.58 & 64.62 & 61.89 & 23.80 \\
DSN-DDI \cite{DSN-DDI}    & 66.49 & 72.46 & 69.97 & 66.49 & 33.05 \\
DeepDDI \cite{DeepDDI-Dataset}    & 72.84 & 66.71 & 77.62 & 72.84 & 36.61 \\
MDDI-SCL \cite{MDDI-SCL}   & 72.93 & 83.25 & 67.41 & 72.93 & 38.02 \\
MDF-SA-DDI \cite{MDF-SA-DDI} & 72.96 & 83.23 & 67.49 & 72.96 & 37.81 \\
HTCL-DDI \cite{HTCL-DDI}   & 74.33 & 53.58 & 88.33 & 74.33 & 10.12 \\
DDIMDL \cite{DDIMDL}    & 82.80 & 94.92 & 90.59 & 82.80 & 54.49 \\
MR-GNN \cite{MR-GNN}     & 83.43 & 87.57 & 86.19 & 83.43 & 57.36 \\
MathEagle \cite{MathEagle}  & 85.85 & 97.01 & 88.51 & 85.85 & 79.70 \\
Taco-DDI \cite{Taco-DDI} & 91.51 & 98.08 & - & 91.47 & - \\

\specialrule{1.2pt}{2pt}{2pt}

GenRel-DDI ($C(*){+}M$)   & \best{93.00} & \best{98.61} & \best{95.24} & \best{93.00} & \best{86.90} \\
GenRel-DDI ($C{+}M(*)$)   & \second{92.92} & \second{98.58} & \second{95.23} & \second{92.92} & \second{86.77} \\
\bottomrule
\end{tabular}%
}
\caption{
Results on DDInter severity, 4-class, reported with micro-averaged ACC, AUROC, AUPR, F1, and MCC. A dash indicates an unreported metric. Pink and gray mark the best and second-best value per metric among reported numbers, with ties highlighted consistently.
}
\vspace{-4mm}
\label{tab:ddinter_singlecol}
\end{table}

\paragraph{Baselines.}
For DeepDDI, ZhangDDI, and ChChDDI, we compare against MR-GNN \cite{MR-GNN}, EPGNN-DS \cite{EPGNN-DS}, DeepDrug \cite{DeepDrug}, MIRACLE \cite{MIRACLE}, CSGNN \cite{CSGNN}, SSI-DDI \cite{SSI-DDI}, DeepDDS \cite{DeepDDS}, DSN-DDI \cite{DSN-DDI}, SMG-DDI \cite{SMG-DDI}, GoGNN \cite{Gognn}, MDF-SA-DDI \cite{MDF-SA-DDI}, CGIB \cite{CGIB}, and IIB-DDI \cite{IIB-DDI}.
For DDInter severity, we compare against MR-GNN \cite{MR-GNN}, GraphSAGE \cite{GraphSAGE}, SSI-DDI \cite{SSI-DDI}, DSN-DDI \cite{DSN-DDI}, DeepDDI \cite{DeepDDI-Dataset}, MDDI-SCL \cite{MDDI-SCL}, MDF-SA-DDI \cite{MDF-SA-DDI}, HTCL-DDI \cite{HTCL-DDI}, DDIMDL \cite{DDIMDL}, MathEagle \cite{MathEagle}, and Taco-DDI \cite{Taco-DDI}.
Baseline numbers are taken from the original papers.

\paragraph{Results.}
Table~\ref{tab:three_datasets_single_table} reports DeepDDI, ZhangDDI, and ChChDDI.
Table~\ref{tab:ddinter_singlecol} reports DDInter severity.
Across tasks, both anchor--adapter variants rank near the top.
This suggests that the same relation trunk transfers beyond the MeTDDI label definition.
On DeepDDI, $C{+}M(*)$ gives the best AUROC, while $C(*){+}M$ gives the best AUPR and ACC.
F1 remains competitive for both variants.
On ZhangDDI, $C(*){+}M$ ranks first among reported methods on AUROC, AUPR, F1, and ACC.
On ChChDDI, $C(*){+}M$ achieves the best AUROC, AUPR, and ACC, while F1 is competitive.
On DDInter severity, $C(*){+}M$ achieves the best scores on all reported metrics, and $C{+}M(*)$ is consistently second.
For binary benchmarks, AUPR is the most informative metric under imbalance.
Several baselines do not report AUPR, but GenRel-DDI achieves the best reported AUPR on both ZhangDDI and ChChDDI.

\section{Conclusion}
Entity-disjoint generalization in DDI prediction remained difficult even with larger single-molecule encoders.
Across overlap-controlled splits, scaling helped on overlap-permitting settings but left a persistent gap on S3.
This suggests that single-molecule objectives do not capture the relation-level patterns needed for unseen drugs under sparse and imbalanced supervision.
GenRel-DDI was introduced to reframe DDI prediction as relation learning, where interaction representations are learned independently of drug identities.
A fixed pretrained encoder provides stable molecular semantics, while a lightweight trainable relation module captures partner-dependent effects.
Across seven public benchmarks with diverse label spaces, GenRel-DDI achieved state-of-the-art or best reported performance under the hardest settings, including strict entity-disjoint evaluation and cross-dataset zero-shot transfer.
Moreover, under overlap restriction, role assignment and freezing choices affected performance more than encoder scale alone.
This supports using a fixed pretrained encoder with targeted adaptation for generalization to unseen drugs.

\newpage
\bibliographystyle{named}
\bibliography{ijcai26}

\end{document}